\documentclass[journal]{IEEEtran}
\usepackage{latexsym,amssymb,amsmath,graphicx,epstopdf,epsf,epsfig,cite,bbm,float,graphics}
\usepackage{algpseudocode}
\usepackage{algorithm}
\usepackage{algorithmicx}
\floatname{algorithm}{Algorithm}
\renewcommand{\algorithmicrequire}{\textbf{Input:}}
\renewcommand{\algorithmicensure}{\textbf{Return:}}

\ifCLASSINFOpdf
\else
\fi

\AtBeginDocument{
\addtolength{\abovedisplayskip}{-1.1ex}
\addtolength{\abovedisplayshortskip}{-1.1ex}
\addtolength{\belowdisplayskip}{-1.1ex}
\addtolength{\belowdisplayshortskip}{-1.1ex}
\addtolength{\belowcaptionskip}{-4ex}
}

\def\ninept{\def\baselinestretch{0.965}}
\ninept

\newcommand{\be}{\begin{equation}}
\newcommand{\ee}{\end{equation}}
\newcommand{\bea}{\begin{eqnarray}}
\newcommand{\eea}{\end{eqnarray}}

\newcommand{\MB}{\left[\begin{array}}
\newcommand{\ME}{\end{array}\right]}

\newcommand{\ei}{\end{itemize}}
\newcommand{\bi}{\begin{itemize}}

\begin{document}
\title{Data Imputation through the Identification of Local Anomalies}
\author{Huseyin~Ozkan, Ozgun S. Pelvan
        and~Suleyman S.~Kozat,~\IEEEmembership{Senior~Member,~IEEE}

\thanks{ Huseyin Ozkan and Suleyman S. Kozat are with the Department
of Electrical and Electronics Engineering, Bilkent University, Ankara, Turkey. E-mail: \{huseyin,kozat\}@ee.bilkent.edu.tr. Huseyin Ozkan is also with the Department
of Image Processing, MGEO, Aselsan, Ankara, Turkey.}
\thanks{ Ozgun S. Pelvan is with the Department
of Electrical and Electronics Engineering, Middle East Technical University,
Ankara, Turkey.}
\thanks{This work was supported in part by the Turkish Academy of Sciences
 Outstanding Researcher Program and in part by TUBITAK
 under Contract 112E161 and Contract 113E517.}}

\maketitle

\begin{abstract}
\linespread{1}\selectfont
We introduce a comprehensive and statistical framework in a model free setting for a complete treatment of localized data corruptions due to severe noise sources, e.g., an occluder in the case of a visual recording. Within this framework, we propose i) a novel algorithm to efficiently separate, i.e., detect and localize, possible corruptions from a given suspicious data instance and ii) a \emph{Maximum A Posteriori} (MAP) estimator to impute the corrupted data. As a generalization to Euclidean distance, we also propose a novel distance measure, which is based on the ranked deviations among the data attributes and empirically shown to be superior in separating the corruptions. Our algorithm first splits the suspicious instance into parts through a binary partitioning tree in the space of data attributes and iteratively tests those parts to detect local anomalies using the nominal statistics extracted from an uncorrupted (clean) reference data set. Once each part is labeled as anomalous vs normal, the corresponding binary patterns over this tree that characterize corruptions are identified and the affected attributes are imputed. Under a certain conditional independency structure assumed for the binary patterns, we analytically show that the false alarm rate of the introduced algorithm in detecting the corruptions is independent of the data and can be directly set without any parameter tuning. The proposed framework is tested over several well-known machine learning data sets with synthetically generated corruptions; and experimentally shown to produce remarkable improvements in terms of classification purposes with strong corruption separation capabilities. Our experiments also indicate that the proposed algorithms outperform the typical approaches and are robust to varying training phase conditions.
\end{abstract}


\IEEEpeerreviewmaketitle
\section{Introduction}
In many applications from a wide variety of fields, the data to be processed can partially (or even almost completely) be affected by severe noise in several phases, e.g., occlusions during a visual recording or packet losses during transmission in a communication channel. The partial, i.e., localized, corruptions resulted in the data due to such problems often severely degrade the performance of the target application; for instance, face recognition or pedestrian detection under occlusion \cite{TNNLSface2, TNNLSface3, TNNLSface1, dollar}. In order to reduce the impact of this adverse effect, we develop a complete and novel framework, which efficiently detects, localizes and imputes corruptions by identifying the local anomalies in a given suspicious data instance. We emphasize that neither the existence nor, if exists, the location of a corruption is known in our framework. Moreover, the proposed algorithms do not assume a model but operate in a data driven manner.

We consider the local corruptions as statistical deviations from the nominal distribution of the uncorrupted (clean) observations. To detect and localize corruptions, i.e., such statistical deviations, we model a corruption as an anomaly due to an external factor (communication failure in a channel or occluder object in an image), which locally overwrites a data instance and moves it outside the support of the nominal distribution. However, corruptions that we consider as examples of anomalies have further specific properties such that (a) the corruptions in an instance are confined to unknown intervals along the data attributes, i.e., localized, and (b) not only a corrupted part but also all of its subparts are anomalous. Thus, a corruption does not provide an anomaly due to an incompatible combination of normal subparts. Based on these properties that accurately model a wide variety of real life applications, we characterize the event of corruption and formulate the corresponding detection/localization as an anomaly detection problem, cf. \cite{highsupport,TNNLSanomaly, S, V, suggested111, suggested222, mmethod2}.

The introduced algorithm applies a series of statistical tests with a pre-specified false alarm rate to the parts of the suspicious instance after extracting the nominal statistics from a reference (training) data set of uncorrupted (clean) observations. As a result, each part is labeled as anomalous/normal and the local anomalies are identified. These parts are generated and organized through a binary tree partitioning of the data attributes, each node of which corresponds to a part of the suspicious instance. Once the nodes (or parts) are labeled as anomalous/normal on this tree, the patterns of corruption are identified using the aforementioned characterization to detect and localize corruptions. We point out that this localization procedure transforms the nominal distribution into a multivariate Bernoulli distribution with a success probability that precisely coincides with the constant false alarm rate of the local anomaly tests. Considering the hierarchy among the binary labels implied by the tree as a directed acyclic graph, the resulting multivariate Bernoulli distribution achieves a certain dependency structure. Under this condition, we derive the false alarm rate of the proposed framework in detecting the corruptions and show that it is a constant rate, namely, no parameter tuning is required even if the data change.

If a corruption is localized, then we impute/replace the affected attributes with the estimates of the underlying unknown true attributes. For this purpose, we additionally develop a novel \emph{Maximum A Posteriori} (MAP) estimator using the ``score function" defined in \cite{V}. Our estimator exploits the local dependencies among the data attributes, where the locality is encoded in the binary partitioning tree. We point out that the implementation of this MAP estimator does not load extra computational cost since it utilizes the outputs of our anomaly detection approach, which are computed prior to the imputation phase. Furthermore, we also propose a novel distance measure named ``ranked Euclidian distance" as a generalization to the standard Euclidean distance, which is used in the course of the labeling of each part as anomalous/normal. The proposed distance measure is compared with the standard Euclidean distance in the experiments and shown to be superior in terms of detecting and localizing corruptions.

We conduct tests over several well-known machine learning data sets \cite{uci, clayton}, which are exposed to severe data corruptions. Our experiments indicate that the proposed framework achieves significant improvements after imputation up to $80 \%$ in terms of the classification purposes and outperforms the typical approaches. The proposed algorithms are also empirically shown to be robust to varying training phase conditions with strong corruption separation capabilities.

\subsection{Related Work}
In this study, the corrupted attributes are considered to be statistically independent with the underlying unobserved true data, i.e., corrupted attributes are of no use in estimation of the uncorrupted counterparts. Hence, if one knows which attributes are corrupted in an instance, then those attributes can readily be treated as missing data, cf. \cite{TNNLSmissing, extraPAMI4, dempster, rubin, little,mmethod1}. For example, classification and clustering with missing data is a well-studied problem in the machine learning literature. The corresponding studies such as \cite{rubin, smolensky, dempster, little, lastref} are related to inference with incomplete data \cite{rubin} and generative models \cite{smolensky}, where Bayesian frameworks \cite{little} are used for inference under missing data conditions. Alternatively, pseudo-likelihood \cite{besag} and dependency network \cite{heckerman} approaches solve data completion problem by learning conditional distributions. In \cite{thesis}, the probability density of the missing data is modeled conditioned on a set of introduced latent variables and thereafter, a MAP based inference is used. However, all of the studies \cite{TNNLSmissing, extraPAMI4, dempster, rubin, little, thesis, besag, heckerman, smolensky, lastref} either assume the knowledge of the location of the missing attributes or impose strong modeling constraints; as opposed to the model free solutions in this paper.

On the other hand, imputation is commonly used as a pre-processing tool \cite{little}. The Mixture of Factor Analyzers \cite{ramani} approach replaces the missing attributes with samples drawn from a parametric density, which models the distribution of the underlying true data. The proposed imputation techniques in \cite{nonparametric1, nonparametric2} are, whereas, both non-parametric and based on the inference of the posterior densities via certain kernel expansions. On the contrary, the MAP estimator in this study does not even attempt to estimate the posterior density either in a parametric or non-parametric manner. Instead, the introduced method is only based on the sufficient rank statistics. We emphasize that unlike our approach, the incomplete data approaches generally assume the knowledge of the missing attributes, i.e., they are precisely localized and provided beforehand. For example, the occluded pixels in the event of occlusion of a target object in an image cannot be known a priori, which requires a detection and localization step. Since the existing studies do not have such a step, an exhaustive list of the occluded pixels as the result of a manual inspection of the missing attributes is required as an input to the algorithms proposed in the corresponding literature. In this regard, our study is the first to jointly handle the issues of detecting/localizing missing attributes, i.e., corruptions, as well as their imputation in one complete framework. Hence, the generic local corruption detection and imputation algorithm of our framework complements the missing data imputation approaches as an additional merit.

Data imputation and completion is also essential in image processing for handling corrupted images, e.g., \cite{extraPAMI2, buades}. Generally, a corrupted image is restored by explicitly learning the image statistics \cite{weiss, lyu} or by using neural networks \cite{gallinari, jain, extraPAMI3}. These denoising studies do not attempt to localize corruptions in an image, but treat them as a noise and filter it out using statistical approaches applied to the image globally. Even though this is a valid approach for image enhancement, an attempt to correct/enhance an image globally in case of only a localized corruption might be even detrimental since the uncorrupted parts are also affected by global operations. Additionally, it is not usually possible to locally impute corrupted portions using denoising approaches. There exist several studies that aim localization as well. Studies such as \cite{dollar, TNNLSface2} indicate that occlusion, as an example of corruption, is a common phenomenon and detrimental in pedestrian detection as well as face recognition applications. In this regard, detection of occluded, i.e., corrupted, visual objects had been previously investigated in a number of studies \cite{toh, yung, pang, zhang}. In these studies, occlusion detection is performed using domain specific knowledge (visual cues) or external information (object geometry), which, however, are not always available in general data imputation setting. From the machine learning perspective, descriptors are extracted from various parts of the occluded object in \cite{mohan} and similarly; part-based descriptors are weighted with the occlusion measure in \cite{enzweiler} to relieve the corresponding degrading effects. Since these approaches do not directly target handling occlusions, i.e., corruptions, they only provide partial or limited solutions. Several other studies propose solutions via extracting occlusion maps, e.g., \cite{wang, mei}. In \cite{wang}, HOG based classification errors; and in \cite{mei}, template based reconstruction errors are used to generate such an occlusion map. However, both studies assume rigid models and significantly rely on domain specific knowledge; and in general fail to remain applicable if the data source belongs to another domain. In this study, we assume that data is generic and no domain information is available, yet detection and imputation of corruption is necessary for improving the subsequent processing stages, such as classification.

\subsection{Summary of Contributions}
\begin{enumerate}
\item This study is the first to jointly handle localized data corruptions in one statistical framework that is designed completely model free for the goal of separating a corruption and imputing the affected data attributes. We also provide a false alarm rate (in detecting corruptions) analysis of the framework via directed acyclic graphs.
\item A novel MAP estimator for imputation and a novel distance measure for corruption localization purposes is proposed.
\item The proposed framework is computationally efficient in the sense that (i) it effectively utilizes a binary search for corruption separation, and (ii) the computational load due to our MAP based imputation is insignificant.
\item We propose a characterization for anomalies, e.g., rarities, incompatible combinations and corruptions, which is a novel notion.\\
\end{enumerate}

In Section \ref{sec:PD}, we provide the problem description. We then present our algorithm in Section \ref{sec:thethemethod} and the associated computational complexity in Section \ref{sec:complexity}. We report the corruption detection/localization performance of the proposed algorithm as well as the improvement in classification tasks achieved by the imputation in Section \ref{sec:experiments}. The paper concludes with a discussion in Section \ref{sec:conclusion}.

\section{Problem Description} \label{sec:PD}
We have a possibly corrupted test instance $\mathbf x \in \mathbb{R}^d$ along with a set of uncorrupted (clean) independent and identically distributed observations $S=\{\mathbf s_1, \mathbf s_2, ..., \mathbf s_{N_s}\}$ as the nominal training (reference) data, where $\mathbf s_i=[{s_i}_1,{s_i}_2,...,{s_i}_d] \in \mathbb{R}^d \sim f_{0}(\mathbf s)$, $d$ is data dimensionality and $f_0$ is the unknown nominal density. The test instance $\mathbf x$ is considered to be corrupted with probability $\pi$ by severe noise in multiple non-overlapping intervals along its dimensions (attributes), which are completely unknown. Suppose that for such an interval, the corruption is localized and confined to the attributes $\mathbf x_{c}^{c+\beta-1}=\{x_c,x_{c+1},...,x_{c+\beta-1}\}$ for some $c$ and $\beta$ in $[1,d]$ with $c+\beta -1 \leq d$. We assume that the corrupted attributes are uniformly and independently distributed, $z_i\in \mathbf x_{c}^{c+\beta-1} \sim U_Z(z)$, where $U_Z$ is the uniform distribution defined in a finite support. Moreover, $Z$ is also statistically independent with the true data and hence, the knowledge of $\mathbf x_{c}^{c+\beta-1}$ is irrelevant to the uncorrupted counterparts. Note that this corruption model implies a total erasure of data in several unknown portions due to an independent source overwriting the attributes in those portions, e.g., an occluder in computer vision applications \cite{TNNLSface2, dollar}. Typically, since no information is provided about the independent source in such applications, we consider that the uniformity assumption draws a worst case scenario and it is realistic.  On the other hand, $\mathbf x$ is considered to be uncorrupted with probability $1-\pi$. Therefore, whether a test instance $\mathbf x$ includes a corruption is unknown; and it is generally modeled to be drawn from the mixture $\mathbf x \sim (1-\pi)f_{0}(\mathbf x) + \pi f_{1}(\mathbf x)$ \cite{V}, where $f_1$ is the probability density of the corrupted instances.

The density $f_1$ can be derived from the unknown nominal density $f_0$ using the described corruption model, if the distributions of $c$, $\beta$ and the number of corrupted intervals are further specified; which is unnecessary in the context of this paper. Hypothetically, if one can correct an instance $\mathbf x$ drawn from the density $f_1$ by replacing all the corrupted attributes, e.g., $\mathbf x_{c}^{c+\beta-1}$, with the underlying true attributes, e.g., $\bar{\mathbf x}_{c}^{c+\beta-1}$, and obtain $\hat{\mathbf x}$, then $\hat{\mathbf x}$ should follow the nominal density $f_{0}$. Similarly, if the corruptions in $\mathbf x$ can be localized, then the corresponding portions would follow the multivariate uniform density $U_{\mathbf Z}(\mathbf z)$ of the appropriate dimensionality. On the other hand, this corruption model potentially creates significant statistical deviations from the reference data since a corrupted observation $\mathbf x \sim f_{1}$ and $f_{1}$, in general, increasingly diverges from $f_{0}$ as the corruption strength increases. Here, the corruption strength can be considered as the number of corrupted attributes and/or the variance of the corruption $U_Z(z)$ that overwrites the true data. Furthermore, our modeling of corruptions poses a missing (incomplete) data problem since the unknown true attributes $\bar{\mathbf x}_{c}^{c+\beta-1}$ in a corrupted interval are statistically irrelevant to the corrupted attributes $\mathbf x_{c}^{c+\beta-1}$. In this paper, by exploiting the statistical deviations from the nominal distribution of observations, we aim to detect and localize the possible corruptions in a given instance $\mathbf x$ and impute the corrupted or missing attributes.

To this end, we formulate an anomaly detection approach to define this framework in Section \ref{sec:thethemethod}, where we draw the distinctions among several examples of anomalous observations and separate the event of corruption. Then, we propose our algorithm and analyze the associated false alarm probability in detecting corruptions as well as the computational complexity.

\section{A Novel Framework for Corruption Detection, Localization and Imputation} \label{sec:thethemethod}
In this section, we develop a novel framework for a complete treatment of possible corruptions in the input data $\mathbf x$. For presentational clarity and without loss of generality, we assume that the input data $\mathbf x$ can be corrupted only in a single interval throughout this section. Note that the generalization to the case of corruptions spread onto several intervals is immediate and indeed, we present a corresponding detailed experiment in Section \ref{sec:experiments}. Since the corruptions are modeled as local statistical deviations within this framework, we give a brief description of the anomaly detection approach that we work with in Section \ref{sec:corruptionsandanomalies}. Based on the characterization of corruptions through their distinctive properties in Section \ref{sec:treelozalization}, we present Algorithm TCS (Tree-based Corruption Separation). After we derive a novel MAP estimator for imputation in Section \ref{sec:theimp}, we derive the false alarm rate of the proposed framework in detecting the corruptions in Section \ref{sec:ACF}.

\vspace{-0.1 in}
\subsection{Detection of Statistical Deviations: Anomalies} \label{sec:corruptionsandanomalies}
A localized corruption is considered to affect an instance in a certain part(s) such that the affected attributes statistically deviate from the vast majority of the data. The proposed algorithm in this paper localizes the corrupted attributes by identifying the local anomalies through a series of statistical checks of the test instance with the reference data. In this section, we briefly describe the anomaly detection approach that we work with and present a novel distance measure for the corruption localization purpose.

The probability density of a possibly corrupted test instance $\mathbf x$ can be modeled as
\begin{align*}
\mathbf x \sim (1-\pi)f_{0}(\mathbf x) + \pi f_{1}(\mathbf x), \text{ where }
\end{align*}
$ H_0: \mathbf x \sim f_0(\mathbf x) $ is the null hypothesis from which the nominal data are drawn,
$ H_1: \mathbf x \sim f_1(\mathbf x) $ is the hypothesis representing the corrupted observations, and $\pi \in [0,1]$ is the corresponding mixing coefficient. Within the framework of anomaly detection approaches, the nominal distribution $f_0$ is usually assumed unknown or hard to estimate; and instead, a set of nominal observations is provided. Then for a given test instance $\mathbf x$, the task in \cite{V} is to decide whether the null hypothesis $H_0$ was realized or the alternative $H_1$ such that the detection rate (of anomalies) is maximized with a constant false alarm rate $\tau$. For this purpose, the score function \cite{V}
\begin{align}
\hat{p}_K(\mathbf x)=\frac{1}{N_s}\sum_{i=1}^{N_s} \mathbf 1_{\{R_S(\mathbf x;K) \leq R_S(\mathbf s_i;K)\}} \label{score}
\end{align}
is proposed, where $\mathbf 1_{\{.\}}$ is the indicator function and $R_S(\mathbf x;K)$ is the Euclidean distance from $\mathbf x$ to its nearest $K$'th neighbor in $S$, if $\mathbf x \notin S$; and to its nearest $K+1$'th neighbor in $S$, otherwise. Based on this score function, the test instance $\mathbf x$ is declared as anomalous \cite{V}, if
\begin{align}
\hat{p}_K(\mathbf x)\leq \tau. \label{test}
\end{align}
When the mixing distribution $f_1$ is assumed uniform, it is shown in \cite{V} that $\hat{p}_K(\mathbf x)$ is an asymptotically consistent estimator of the density level of the test instance
\begin{align}
p(\mathbf x)=\int_{\forall \mathbf s} \mathbf 1_{\{f_0(\mathbf x) \geq f_0(\mathbf s)\}} f_0(\mathbf s) d\mathbf s \label{densitylevel}
\end{align}
under certain smoothness conditions. Remarkably, $\{x: p(x)\geq \tau\}$ provides the minimum volume set at level $\tau$, which is the most powerful decision region for testing $H_0$ vs $H_1$ with a constant false alarm rate $\tau$ \cite{S}. We note that the precision of the test defined in \eqref{test} degrades faster with the dimensionality than it improves with the size of the training data. As a result, we here point out several practical issues about detecting the existence of a corruption with this approach.

Briefly, i) a direct test of an instance $\mathbf x$ does not localize a possible corruption for imputation, ii) on the contrary, a truly corrupted instance, i.e., an instance of hypothesis $H_1$, does not necessarily test positive due to the limited training data, high dimensionality as well as that the corruption might not be sufficiently strong; and iii) corruptions have further specific properties in addition to that they provide anomalies, which must be incorporated to achieve a better false alarm rate compared to $\tau$.\\

\emph{Ranked Euclidean Distances}:
To address the first issue in this list, we propose a novel distance measure (not a metric in the mathematical sense), which is sensitive to only a certain $\alpha$ fraction of the attributes for a given pair of instances $\mathbf x$ and $\mathbf y$. For instance, a corruption of only a single attribute in a given test instance $\mathbf x$ might be significantly strong such that the whole instance turns anomalous with the test in \eqref{test} used with the standard Euclidean distance. In this case, any part of the instance $\mathbf x$ including the corrupted attribute would test positive, which creates an ambiguity in terms of the localization, i.e., separation, of the corrupted attribute, and in turn requires an exhaustive search over all possible subsets in the space of the attributes.

To overcome such ambiguities, we propose a distance measure so that the test in \eqref{test} results positive only when the corruption has a sufficiently large support, which disregards a pre-specified fraction of the attributes that are most responsible for a possible corruption. We define this measure for an $\alpha\in [0,1]$ as
\begin{align}
h_\alpha(\mathbf x, \mathbf y) = \sqrt{\sum_{i=1}^{\lfloor d\alpha \rfloor}(x_{k(i)} - y_{k(i)})^2}, \label{distancemeasure}
\end{align}
where $k$ is a permutation of the attributes with
\begin{align*}
|x_{k(1)} - y_{k(1)}| &\leq ...\leq|x_{k(i)} - y_{k(i)}|\leq...\leq|x_{k(d)} - y_{k(d)}|,
\end{align*}
and $\lfloor . \rfloor$ is the floor operator.
Since this distance measure depends only on the $\alpha$ fraction of the least deviated attributes between $\mathbf x$ and $\mathbf y$, a corruption must have a support of at least $(d-\lfloor d\alpha \rfloor)$-length to make an instance anomalous with respect to the reference data. Here, $(1-\alpha)$ can be seen as the precision of the localization when an anomalous instance is checked with the test in \eqref{test} using the distance measure defined in \eqref{distancemeasure}. This precision obviously cannot be made arbitrarily large since as $1-\alpha$ approaches $1$, the distance $h_\alpha$ becomes more prone to noise and the correlation structure between the attributes is less exploited. We investigate this trade-off further in our simulations. The distance measure $h_\alpha$ recovers the standard Euclidean distance when $\alpha=1$ and will be named in the rest of the paper as ``ranked Euclidean distance". We note that for the cases $\alpha<1$, $h_\alpha$ fails to be a metric in the mathematical sense, i.e., $h_\alpha(x,y)=0 \Leftrightarrow x=y$ is not satisfied, which requires to specify a nominal density model on $f_0$ to derive the same asymptotic consistency in \cite{V} for the score values $\hat{p}_K(x)$ in estimating the density levels $p(x)$ with $h_\alpha$. However, we do not assume -in this work- any density model for $f_0$ or do not take any stochastic assumptions regarding the data source.

In the following section, we characterize the corruptions by presenting their specific properties and propose an algorithm to localize and impute corruptions.

\vspace{-0.1 in}
\subsection{Modeling of Localized Corruptions} \label{sec:treelozalization}
\begin{figure}[t]
\centerline{\epsfxsize=8cm \epsfysize=5cm \epsfbox{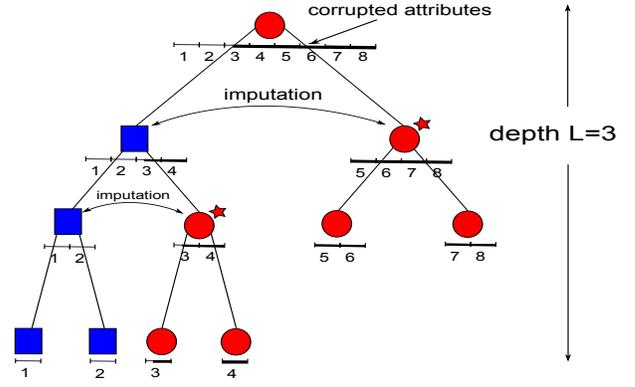}}
\caption{An illustration of Algorithm TCS (Tree-based Corruption Separation) with $\alpha=0.5$}
\label{fig:example1} \vspace{-0.2 in}
\end{figure}
If a test instance is subject to corruption in a small part only, the corruption might not be detectable when it is checked using an anomaly detection algorithm without a detailed analysis in its parts. On the other hand, an anomalous observation does not necessarily contain a corruption since it might simply be a false alarm, in fact an uncorrupted observation. To address these two issues, we propose a statistical analysis of a test instance through its parts using a binary partitioning tree in the space of data attributes on which, we also provide a characterization to separate the event of corruption among possible anomaly scenarios.

Suppose that an instance $\mathbf x = [x_1,x_2,...,x_d]\in \mathbb{R}^d$ corresponds to the root node $R$ on a binary tree. Using half-way splits for presentational simplicity, let the set of attributes $V_{R_l}=\{x_1,x_2,...,x_{\lfloor \frac{d}{2} \rfloor}\}$ be assigned to the left child node $R_l$ of the root and $V_{R_r}=\{x_{{\lfloor \frac{d}{2} \rfloor}+1},x_{{\lfloor \frac{d}{2} \rfloor}+2},...,x_d\}$ assigned to the right child node $R_r$, Fig. \ref{fig:example1}. Note that $V_R=\{x_1,x_2,...,x_d\}$ with $V_{R_l} \cap V_{R_r} = \emptyset$ and $V_{R}=V_{R_l}\cup V_{R_r}$. Based on this strategy for generating subparts of an instance, we propose Algorithm TCS (Tree-based Corruption Separation) to separate and impute corruptions, which recursively expands a depth-$L$ binary tree to partition the space of attributes. For each node $\nu$ created in the course of this expansion, the corresponding attributes/part of the test instance, e.g., $\mathbf x_{V_{R_l}}:=\mathbf x_1^{\lfloor \frac{d}{2} \rfloor}$ with $\nu=R_l$, is checked whether it is consistent with the reference data restricted to those attributes, e.g., $S_{V_{R_l}}=\{{s_1}_1^{\lfloor \frac{d}{2} \rfloor},{s_2}_1^{\lfloor \frac{d}{2} \rfloor},...,{s_{N_s}}_1^{\lfloor \frac{d}{2} \rfloor}\}$ with $\nu=R_l$, using the test defined in \eqref{test}. We here use the ranked Euclidean distance $h_{\alpha}$ in this testing with a pre-specified $\alpha$. Therefore, each node $\nu$ encountered in this expansion is assigned a binary label as anomalous/normal and a fully labeled (possibly unbalanced) tree is obtained for the test instance $\mathbf x$. We emphasize that Algorithm TCS does not completely construct this depth $L$-binary tree at the beginning but instead expands it by creating the nodes and the edges as needed to achieve an efficient implementation, which continues until that each data attribute is decided to be corrupted or uncorrupted.

\begin{figure}[t]
\centerline{\epsfxsize=8cm \epsfysize=3.5cm \epsfbox{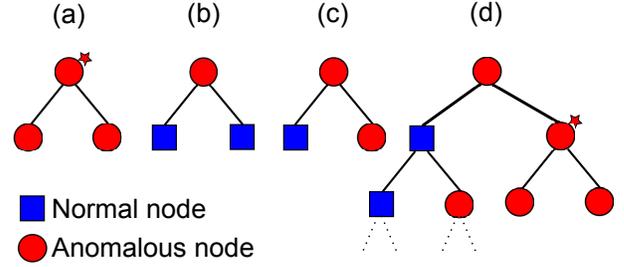}}
\caption{An anomalous observation with several scenarios in its parts. Note that the starred nodes indicate localized corruptions. (a) A conclusive pattern: corruption is detected. (b) A conclusive pattern: corruption is rejected. (c) An inconclusive pattern: Anomaly consisting of normal and anomalous parts needs to be further explored. (d) Further exploration of the test instance to locate a possible corruption by searching a conclusive pattern.}
\label{fig: simplecorruption} \vspace{-0.2 in}
\end{figure}

We consider several scenarios where the observation $\mathbf x_{V_{\nu}}$ at a node $\nu$ can be anomalous.
In Fig. \ref{fig: simplecorruption}, the nodes are illustrated as circles, if the corresponding part is found to be anomalous; and squares otherwise.  An anomaly can be wide-spread onto the attributes and consist of anomalous subparts as illustrated in Fig. \ref{fig: simplecorruption}a, which is regarded as a conclusive pattern since a corruption is characterized and defined in Section \ref{sec:PD} by that all of the subparts of a corrupted instance are also corrupted. Hence, a corruption at the starred node in Fig. \ref{fig: simplecorruption}a is declared, unless it is the root node. Note that a global corruption at the root is disregarded in this work since it is not localized. In another case, an anomalous observation could be non anomalous in its parts as illustrated in Fig. \ref{fig: simplecorruption}b, which simply happens due to an incompatible or rare combination of attributes in its subparts. This is a typical situation, where an anomalous observation is not corrupted. Hence, this case also provides a conclusive pattern in our consideration such that a corruption is rejected at the anomalous node. On the contrary, the case in Fig. \ref{fig: simplecorruption}c is an inconclusive pattern, which suggests a corruption at the right child, however, whether the corruption is spread in attributes of that child or localized is unknown. Hence, the attributes of the right child is further split and explored similarly. Then, if the conclusive pattern in Fig. \ref{fig: simplecorruption}a (or Fig. \ref{fig: simplecorruption}b) is realized, then the corruption is accepted and localized (or rejected) at the starred node in Fig. \ref{fig: simplecorruption}d. Otherwise, the search continues. On the other hand, if a significantly small subset of the corrupted attributes are left at the left child node in Fig. \ref{fig: simplecorruption}c, it might not be detectable and labeled as normal. Then the corresponding attributes should further be split as illustrated in Fig. \ref{fig: simplecorruption}d. This process recursively defines a corruption localization with an improved false alarm rate as several anomalies are rejected as they are false alarms, i.e., non corrupted anomalies.

The introduced Algorithm TCS then searches in a breadth-first-search fashion the described binary tree for a corruption. When the conclusive (or terminating) pattern shown in Fig. \ref{fig: simplecorruption}a (Fig. \ref{fig: simplecorruption}b) is found in the course of this expansion, the search is stopped at the parent node of the found pattern, i.e., the tree is pruned on that branch, and corruption is declared (or no corruption is found and no action is necessary) for the corresponding attributes. This search of corruption at each branch starting from the root node continues to the corresponding leaf node, unless a terminating pattern is found. Finally, if a conclusive pattern is not encountered at a branch from the root to an anomalous leaf, we opt to accept the corruption at the leaf to favor a better detection at a cost of an increased corruption false alarm rate. An illustration of the progress of the algorithm is given in Fig. \ref{fig:example1}, where the corrupted attributes are successfully located. Note that a small set of the attributes are mislabeled as corrupted, i.e., false alarms, in the region 3, which can be corrected if the partitioning resolution is improved by increasing the depth $L$.

\vspace{-0.1 in}
\subsection{\emph{Maximum A Posteriori} (MAP) Based Imputation} \label{sec:theimp}
We emphasize that in most of the detection and estimation applications, the posterior density, e.g., $f_0(\bar {\mathbf x}_{V_\nu}|\mathbf x)$ in \eqref{ref1}, of the target is too complicated to assume realistic parametric models so that the nonparametric approaches are often favored in such situations \cite{ML}. In accordance, we introduce an algorithm that works under a completely model free setting regarding both the localization of the corruptions and the imputation. Furthermore, we point out that MAP based estimators are generally known to generate more plausible results when the posterior density is multi-modal, compared to MMSE based estimators, i.e., simple (possibly weighted) averaging, which can even generate infeasible solutions \cite{MAPvsMMSE1,MAPvsMMSE2,MAPvsMMSE3}. This is often the case especially for the computer vision and machine learning applications such as edge preserving image denoising \cite{MAPvsMMSE4}. For instance, the gradients in an occluded pedestrian image would get too smoothed in an MMSE based imputation, which might cause the gradient based feature extractors, e.g., HOG \cite{hog}, to fail or not perform satisfactory in case of an pedestrian detection application \cite{dollar, ML}. For these reasons, we propose a novel MAP based imputation technique, which always generates feasible and likely estimates and approximates the true MAP estimator as the size of the reference data increases.

Once a corruption is localized for an instance $\mathbf x$ at a node $\nu$, then our task is to estimate the original attributes $\bar{\mathbf x}_{V_\nu}$ using the training data set $S$ as well as the instance $\mathbf x$ and impute accordingly, i.e., replace the corrupted attributes in $\mathbf x$ with the estimates. Since we assume the corrupted attributes $\mathbf x_{V_\nu}$ to be statistically independent with the underlying true data $\bar{\mathbf x}_{V_\nu}$, we treat the corrupted attributes as missing data, which then should have no effect in estimation of the true attributes. Hence, we condition this estimation of the data $\bar{\mathbf x}_{V_\nu}$ on the remaining attributes in $\mathbf x$. On the other hand, we note that in most of the applications such as the image compression \cite{spatial}, the data attributes being in sufficiently close proximity are usually modeled to manifest high correlation. In accordance, we propose to estimate the unknown data $\bar{\mathbf x}_{V_\nu}$ conditioned on the attributes $\mathbf x_{V_{\nu_s}}$ associated with its nearest neighbor on our tree, i.e., the sibling node $\nu_s$ of $\nu$. Note that due to the localization of corruptions by Algorithm TCS (Tree-based Corruption Separation), the attributes at the sibling node $\nu_s$ are certainly detected to be uncorrupted in case of the standard Euclidean distance; and detected to be uncorrupted with significantly high probability in case of the ranked Euclidean distance (cf. Section \ref{sec:ACF}). In the following, we introduce a novel \emph{Maximum A Posteriori} (MAP) estimator of the true data underlying the corrupted attributes based on the standard Euclidean distance ($h_{\alpha}$ with $\alpha=1$) and then discuss the generalization over $\alpha$ for the ranked Euclidean distance measure. We also stress that the implementation of this estimator is only based on the outputs of our corruption localization algorithm, which are computed before the imputation phase in the course of Algorithm TCS. Therefore, computationally, the imputation phase that we develop is efficient such that it does not require further computations.
\begin{algorithm}[t]
\caption{TCS --- Tree-based Corruption Separation}
\label{alg:desc}
\algorithmicrequire \text{} $\alpha,K,\tau,L$; \text{} $S,\mathbf x$
\begin{algorithmic}[1]
\State Initialize $\mathcal{C}\leftarrow\emptyset$: set of corrupted attributes
\State Initialize $\mathbf y \leftarrow \mathbf x$: imputed test data
\State Create the root node $\nu \leftarrow R$ and label
\Procedure{recurse}{$\nu$}
\State Create nodes $\nu_{l}$ and $\nu_{r}$; and label
\If{the pattern in Fig. \ref{fig: simplecorruption}a}
\If{$\nu$ is the root} \Return
\Else
\State Declare corruption at $\nu$: $\mathcal{C}\leftarrow\mathcal{C}\cup V_{\nu}$
\State Impute attributes $V_{\nu}$ in $\mathbf y$
\State\Return
\EndIf
\ElsIf{the pattern in Fig. \ref{fig: simplecorruption}b} \Return
\ElsIf{$\nu$ is a parent of a leaf}
\If{$\nu_j$ ($j=l$ or $j=r$) is anomalous}
\State Declare corruption at $\nu_j$: $\mathcal{C}\leftarrow\mathcal{C}\cup V_{\nu_j}$
\State Impute attributes $V_{\nu_j}$ in $\mathbf y$
\EndIf
\State\Return
\Else
\State{\Call{recurse}{$\nu_l$}} and {\Call{recurse}{$\nu_r$}}
\EndIf
\EndProcedure
\end{algorithmic}
\algorithmicensure \text{ }$\mathcal{C}$ and $\mathbf y$
\end{algorithm}

Since the only relevant part of the test instance $\mathbf x$ to the proposed MAP estimator is $\mathbf x_{V_{\nu_s}}$, we have
\begin{align}
f_0(\bar {\mathbf x}_{V_\nu}|\mathbf x)=f_0(\bar {\mathbf x}_{V_\nu}|\mathbf x_{V_{\nu_s}}), \label{ref1}
\end{align}
where $\bar {\mathbf x}_{V_\nu}$ represents a realization of the conditional probability density of the true data underlying the corrupted attributes $V_{\nu}$. Then, the MAP estimator of $\bar {\mathbf x}_{V_\nu}$ maximizes the posterior distribution as
\begin{align*}
\mathbf x^{MAP}_{V_\nu}&=\arg\sup_{\bar {\mathbf x}_{V_\nu}\in \mathbb{R}^{|V_\nu|}}f_0(\bar {\mathbf x}_{V_\nu}|\mathbf x_{V_{\nu_s}}).
\end{align*}
For any $\epsilon > 0$, and under certain smoothness constraints on $f_0$ with $f_0(\bar{\mathbf x}_{V_\nu})\neq 0$, let
\begin{align*}
B_\epsilon(\bar {\mathbf x}_{V_\nu}) \cap S_{V_\nu} \neq \emptyset
\end{align*}
hold with some probability $\delta_{N_s}$, where $B_\epsilon(\bar {\mathbf x}_{V_\nu})$ (w.r.t. the standard Euclidean distance) is the $\epsilon$-ball around $\bar {\mathbf x}_{V_\nu}$ in $\mathbb{R}^{|V_\nu|}$ and $N_s=|S|$. Then we point out that
\begin{align*}
\lim_{N_s\rightarrow\infty} \delta_{N_s}=1.
\end{align*}
Hence, since $\epsilon$ can be made arbitrarily small, we obtain
\begin{align*}
\mathbf x^{MAP}_{V_\nu}&=\arg\lim_{N_s\rightarrow\infty}\sup_{\bar {\mathbf x}_{V_\nu}\in S_{V_\nu}}f_0(\bar {\mathbf x}_{V_\nu}|\mathbf x_{V_{\nu_s}}),
\end{align*}
and by the Bayes' rule
\begin{align}
\mathbf x^{MAP}_{V_\nu}&=\arg\lim_{N_s\rightarrow\infty}\sup_{\bar {\mathbf x}_{V_\nu}\in S_{V_\nu}} \frac{f_0(\bar {\mathbf x}_{V_\nu},\mathbf x_{V_{\nu_s}})}{f_0(\mathbf x_{V_{\nu_s}})} \nonumber\\
&=\arg\lim_{N_s\rightarrow\infty}\sup_{\bar {\mathbf x}_{V_\nu}\in S_{V_\nu}} f_0(\bar {\mathbf x}_{V_\nu},\mathbf x_{V_{\nu_s}})\label{MAP}
\end{align}
with probability $1$, where the denominator is dropped since it does not depend on the maximizer, i.e., $\bar {\mathbf x}_{V_\nu}$. To approximate the MAP estimator given in \eqref{MAP}, we adapt the nonparametric k-nearest neighbor (knn) based density estimation approach \cite{silverdensity}. Let us define a small neighborhood around $\mathbf x_{V_{\nu_s}}$ in $\mathbb{R}^{|V_{\nu_s}|}$ as
\begin{align}
\mathcal{N}_{N_s}(\mathbf x_{V_{\nu_s}})=\{\mathbf s: R_S(x_{V_{\nu_s}};{\gamma \sqrt{N_s}})\geq h_{\alpha=1}(\mathbf x_{V_{\nu_s}}, \mathbf s)\} \label{empiricalneighborhood},
\end{align}
\noindent where $h_{\alpha=1}(.,.)$ is the Euclidean distance and $R_S(x_{V_{\nu_s}};{\gamma \sqrt{N_s}})$ is the $h_{\alpha=1}(.,.)$ distance from $x_{V_{\nu_s}}$ to its nearest $\gamma \sqrt{N_s}$'th neighbor in $S_{V_{\nu_s}}$ for some $\gamma>0$. Note that as $N_s \rightarrow \infty$, $\mathcal{L}(\mathcal{N}_{N_s}(\mathbf x_{V_{\nu_s}}))\rightarrow 0$, where $\mathcal{L}(.)$ is the Lebesgue measure. Then \eqref{MAP} yields
{\small
\begin{align}
\mathbf x^{MAP}_{V_\nu}=\arg\lim_{N_s\rightarrow\infty}\sup_{\bar {\mathbf x}_{V_\nu}\in S_{V_\nu}} \frac{\int_{\mathbf z\in \mathcal{N}_{N_s}(\mathbf x_{V_{\nu_s}})}f_0(\bar {\mathbf x}_{V_\nu},\mathbf z)\mathbf d \mathbf z}{\mathcal{L}(\mathcal{N}_{N_s}(\mathbf x_{V_{\nu_s}}))} \label{MAP2},
\end{align}}
\noindent with probability $1$. When $N_s$ is sufficiently large with $N_s\geq N_s^*$ for some $N_s^*$ or $\mathcal{L}(\mathcal{N}_{N_s})$ is sufficiently small, we assume that $f_0(\bar {\mathbf x}_{V_\nu},\mathbf x_{V_{\nu_s}})$ is subject to negligible variations only. Then, we (with probability $1$) obtain the approximation:
\begin{align}
\mathbf x^{MAP}_{V_\nu}&=\arg\lim_{N_s\rightarrow\infty}\sup_{\bar {\mathbf x}_{V_\nu}\in S_{V_\nu}} \frac{\int_{\mathbf z\in \mathcal{N}_{N_s}(\mathbf x_{V_{\nu_s}})}f_0(\bar {\mathbf x}_{V_\nu},\mathbf z)\mathbf d \mathbf z}{\mathcal{L}(\mathcal{N}_{N_s}(\mathbf x_{V_{\nu_s}}))}  \nonumber\\
&\simeq\arg\max_{\bar {\mathbf x}_{V_\nu}\in S_{V_\nu},\mathbf z\in \mathcal{N}_{N_s^*}(\mathbf x_{V_{\nu_s}})} f_0(\bar {\mathbf x}_{V_\nu},\mathbf z)\label{MAP3},
\end{align}
where, in order to obtain the corresponding maximum in the reference set $S$, knowing the rank statistics in $f_0(\bar {\mathbf x}_{V_\nu},\mathbf z)$ is enough, i.e., explicitly estimating/computing the density is unnecessary. Therefore, using the density function defined in \eqref{densitylevel}, we obtain
\begin{align}
\mathbf x^{MAP}_{V_\nu}\simeq\arg\max_{\bar {\mathbf x}_{V_\nu}\in S_{V_\nu},\mathbf z\in \mathcal{N}_{N_s^*}(\mathbf x_{V_{\nu_s}})} p(\bar {\mathbf x}_{V_\nu},\mathbf z). \label{usethis1}
\end{align}
For sufficiently large $N_s$, note that $\hat{p}_K(\bar {\mathbf x}_{V_\nu},\mathbf z)$ approximates $p(\bar {\mathbf x}_{V_\nu},\mathbf z)$ \cite{V}, i.e., $\forall (\bar {\mathbf x}_{V_\nu},\mathbf z)$
\begin{align}
|\hat{p}_K(\bar {\mathbf x}_{V_\nu},\mathbf z)-p(\bar {\mathbf x}_{V_\nu},\mathbf z)| \simeq 0 \text{ almost surely }. \label{usethis2}
\end{align}
Using the result in \eqref{usethis1} in combination with \eqref{usethis2}, we propose to use MAP based estimator of the true data underlying the corrupted attributes
\begin{align}
\mathbf x^{MAP}_{V_\nu}\simeq\hat{\mathbf x}_{V_\nu}=\arg\max_{\bar {\mathbf x}_{V_\nu}\in S_{V_\nu},\mathbf z\in \mathcal{N}_{N_s^*}(\mathbf x_{V_{\nu_s}})}\hat{p}_K(\bar {\mathbf x}_{V_\nu},\mathbf z),\label{theestimator}
\end{align}
based on which we replace, i.e., impute, the corrupted attributes $\mathbf x_{V_\nu}$ in the instance $\mathbf x$ with $\hat{\mathbf x}_{V_\nu}$ and obtain the imputed data as $\mathbf y$.

This estimator is implemented in Algorithm TCS (Tree-based Corruption Separation) at every node in the tree, where a corruption is detected. Namely, we i) obtain the $K$ neighbors of the test instance in the reference data set $S$ with respect to the attributes associated with the node $\nu_s$; ii) for those neighbors in $S$, find the one, say $s^*$, attaining the largest score value defined in \eqref{score} using the attributes associated with the parent node $\nu_p$; then iii) impute the instance $\mathbf x$, which is detected to be corrupted at the node $\nu$, using $s^*$ for the attributes $V_\nu$. In the realistic case of high dimensional and limited data, when the standard Euclidean distance is used as in our derivations, $\mathbf x_{V_{\nu_s}}$ might include corrupted attributes even though it is detected as normal, which clearly adversely affects the calculation of the neighborhood $\mathcal{N}_{N_s}(\mathbf x_{V_{\nu_s}})$ in \eqref{empiricalneighborhood}. In addition, $\mathbf x_{V_{\nu}}$ might only include a small support of corruption, and then we would not like to impute $\mathbf x_{V_{\nu}}$ completely. To overcome these two issues, we propose to use the ranked Euclidean distance defined in \eqref{distancemeasure}. To this end, the neighborhood $\mathcal{N}_{N_s}(\mathbf x_{V_{\nu_s}})$ is defined using $h_\alpha$ with an appropriate $\alpha\neq 1$ in \eqref{empiricalneighborhood}. This cancels the adverse effect, up to a certain degree, of a possible corruption in $\mathbf x_{V_{\nu_s}}$ as desired. Nevertheless, recalling that $h_\alpha$ only uses the $\alpha$ fraction of the attributes $V_{\nu_s}$ and set the others free, $h_{\alpha}$ is not a metric in the mathematical sense and then, as $N_s \rightarrow \infty$, $\mathcal{L}(\mathcal{N}_{N_s}(\mathbf x_{V_{\nu_s}}))\rightarrow 0$ does not hold. As a result, the correlation structure given in \eqref{ref1} is less exploited in imputation as $\alpha$ decreases. Meanwhile, as $\alpha$ decreases, the support of the detected corruption in $x_{V_\nu}$ increases, i.e., localization improves. Therefore, we obviously have a trade-off between the imputation quality and the localization, which is sensitive to the choice of $\alpha$ and investigated in the experiments in greater detail. However, $\alpha$ should be set typically around $0.5 - 0.75$ since we use half-way splits. Finally, note that the imputation brings almost no further computational complexity, since these steps do computationally only depend on the anomaly detection results, cf. \eqref{test} and \eqref{score}, at the corrupted node, its sibling node as well as its parent node, which are all generated prior to the imputation steps.

In the following section, the proposed framework is shown to achieve a constant false alarm rate in terms of the corruption detection. Moreover, this false alarm rate is precisely calculated under a certain dependency structure among the anomalous/normal labels on the partitioning tree.

\vspace{-0.1 in}
\subsection{False Alarm Rate in Detecting Corruptions} \label{sec:ACF}
Since the imputation is an ``overwriting" operation, whether or not to impute a suspicious instance is certainly a ``critical" decision. In case of a false decision if the suspicious instance is in fact uncorrupted, i.e., ``a false alarm in detecting corruptions", the imputation would correspond to data loss. In this section, we study the rate of such occurrences and analyze the false alarm rate of the proposed algorithms in detecting corruptions.

The anomaly detection test applied at every node in Algorithm TCS (Tree-based Corruption Separation) operates with a constant false alarm rate $\tau$, whereas the proposed approach is able to reject corruptions at anomalous nodes. For example, when the terminating pattern in Fig. \ref{fig: simplecorruption}b is encountered, all the anomalies that can be present in the tree rooted from the terminating pattern are rejected, i.e., they are not counted as corruptions. For this reason, the false alarm rate of the proposed approach must be defined in the sense of corruptions as opposed to anomalies. To analyze this false alarm rate in detecting corruptions, one also must account for that the anomaly detection test at a node could be strongly correlated with the outputs of the previous tests in the course of Algorithm TCS, since the data attributes are in general correlated. In this section, we first model the labeling of the nodes, i.e., anomalous vs normal, on the partitioning tree, cf. Fig. \ref{fig:example1}, as a directed acyclic graph \cite{dag} achieving a certain dependency structure and then derive the false alarm rate of Algorithm TCS. Under this modeling, we also show that the constant false alarm rate in detecting the local anomalies at each node globally maps to also a constant false alarm rate in detecting the corruptions.

\begin{figure}[t]
\centerline{\epsfxsize=4cm \epsfysize=3cm \epsfbox{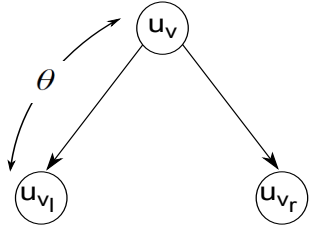}}
\caption{We assume the conditional independency: $p_0(u_\nu,u_{\nu_l},u_{\nu_r})$=$p_0(u_{\nu_l}|u_\nu)p_0(u_{\nu_r}|u_\nu)p_0(u_\nu)$. Moreover, $p_0(u_{\nu_l}|u_\nu)$= $(1-\theta)p_0(u_{\nu_l})+\theta 1_{\{u_{\nu_l}=u_\nu\}}$, where $\theta$ defines the dependency between the parent node and its siblings such that a positive covariance is embedded. Note that $\theta=0$ implies independency.}
\label{fig:dag} \vspace{-0.2 in}
\end{figure}

Recall that Algorithm TCS expands the binary tree in Fig. \ref{fig:example1} for a given uncorrupted test instance $\mathbf s$ and declares a corruption only if the conclusive pattern in Fig. \ref{fig: simplecorruption}a is encountered or a leaf node is found anomalous in the described breadth-first search. In addition to the corruption localization as well as the imputation capabilities of the proposed Algorithm TCS, let us denote the corruption detection in Algorithm TCS by $\mathcal{C(\mathbf s)}=1$, if $\mathbf s$ is detected to be corrupted and $\mathcal{C(\mathbf s)}=0$, otherwise. Then our task is to find the false alarm probability in detecting the corruptions, which is given by
\begin{align}
C_\tau = \int_{\forall \mathbf s} \mathcal{C}(\mathbf s) f_0(\mathbf s) \mathbf d \mathbf s \label{genericfalsealarm},
\end{align}
where $\tau$ is the constant false alarm rate of the detection at each node and $f_0$ is the nominal density. Next, we observe that Algorithm TCS maps every data instance to a binary observation such that the nominal distribution $f_0$ is transformed into a multivariate Bernoulli distribution $p_0$, i.e.,
\begin{align*}
&\mathbb{R}^d \rightarrow B^{2^{L+1}-1} \text{ via }\\
&\mathbf s \rightarrow \mathcal{L}(\mathbf s)=\mathbf u = (u_{R}, u_{R_l}, u_{R_r}, u_{{R_l}_l}, u_{{R_l}_r}, u_{{R_r}_l}, u_{{R_r}_r},...),
\end{align*}
where $B=\{-1,1\}$, $L$ is the depth and $u_R$ is the anomaly decision at the root node such that $u_R=1$, if an anomaly detected; and $u_R=-1$, otherwise. Similarly for the others such as $u_{R_l}$ is the decision at the left hand child of the root and $u_{R_r}$ is the decision at the right hand child. Note that the proposed algorithm does not completely construct the binary tree but expands, i.e., the nodes and the edges are created as needed. Therefore, we do not completely observe the binary vector $\mathbf u$ that an instance $\mathbf s$ maps to, however, we temporarily suppose that all the labels are available for ease of exposition. Once $\mathbf s$ is mapped to $\mathbf u$, since Algorithm TCS declares a corruption based on only the vector of binary labels $\mathbf u$, we equivalently have
\begin{align}
C_\tau &= P\left(\mathcal{C}(\mathbf s)=1\text{ }|\text{ }\mathbf s \text{, in fact, is uncorrupted}\right)\nonumber\\
       &= \sum_{\mathbf u \in \{-1,1\}^{2^{L+1}-1}} \mathcal{C}(\mathbf u) p_0(\mathbf u)\nonumber\\
       &= 1- \sum_{\mathbf u \in \{-1,1\}^{2^{L+1}-1}} \mathcal{C}^c(\mathbf u) p_0(\mathbf u) \label{myeq1},
\end{align}
where $\mathcal{C}(\mathbf u)$ is the corruption decision (with abuse of notation), $\mathcal{C}^c(\mathbf u)$ is the complement, i.e., $\mathcal{C}^c(\mathbf u)=1-\mathcal{C}(\mathbf u)$ and $p_0$ is the corresponding nominal probability mass function such that
\begin{align*}
p_0(\mathbf u) = \int_{\forall \mathbf s: \mathcal{L}(\mathbf s)=\mathbf u }f_0(\mathbf s) \mathbf d \mathbf s.
\end{align*}
In order to calculate the probability mass function $p_0$, we model the binary tree, where each node corresponds to a binary random variable, as a directed acyclic graph \cite{dag} such that the binary random variables at any two sibling nodes are independent conditioned on the knowledge of the label at the parent node. Namely, for any non leaf node $\nu$ and its children $\nu_l$ and $\nu_r$ on the binary partitioning tree, we assume the following conditional independency for the associated random labels: $p_0(u_{\nu_l},u_{\nu_r}|u_\nu)= p_0(u_{\nu_l}|u_\nu)p_0(u_{\nu_r}|u_\nu)$, from which we obtain
\begin{align}
p_0(u_\nu,u_{\nu_l},u_{\nu_r})&=p_0(u_{\nu_l},u_{\nu_r}|u_\nu)p_0(u_\nu) \nonumber\\
&=p_0(u_{\nu_l}|u_\nu)p_0(u_{\nu_r}|u_\nu)p_0(u_\nu), \label{condindep}
\end{align}
cf. Fig. \ref{fig:dag}.
Here, we emphasize that $\mathbf s$ (or $\mathbf u$) is assumed to be uncorrupted in the false alarm analysis to calculate the probability given in \eqref{genericfalsealarm}, i.e., it does not have any localized corruptions by definition. Then, if $\mathbf s$ is declared, at the root node without loss of generality, as anomalous then this anomaly is not due to a corruption but simply a ``rarity" as the test in \eqref{test} is based on density levels. On the contrary to the case of corruption, since a ``rarity" at a node is not a localized phenomenon, we expect that the children inherit the parent label independently. Therefore, we assumed the conditional independency in \eqref{condindep} as a generating dependency structure for the simplest graph presented in Fig. \ref{fig:dag}, which straightforwardly generalizes to the binary tree of the anomalous vs normal labels from root to the leaves. Based on this, we obtain
\begin{align}
&p_0(\mathbf u) = p_0(\mathbf u_R|u_R)^*p_0(u_R) \nonumber\\
&=p_0(\mathbf u_{R_l} \mathbf u_{R_r}|u_R, u_{R_l}, u_{R_r})p_0(u_{R_l}, u_{R_r}|u_R)p_0(u_R) \label{temp1}\\
&=p_0(\mathbf u_{R_l} |u_{R_l})^*p_0(\mathbf u_{R_r} |u_{R_r})^*p_0(u_{R_l}|u_R)p_0(u_{R_r}|u_R)p_0(u_R) \nonumber,
\end{align}
where $\mathbf u_R$ is the collection of the binary variables associated with the nodes in the tree rooted from node $R$ that excludes $u_R$, and the last equation follows from \eqref{condindep} and the Bayes' rule. We observe that the starred factors in the expression \eqref{temp1} are of similar forms such that the last equation can be expanded further using similar lines of derivations up until the leaves appear.

Thus, the calculation of $p_0(\mathbf u)$ requires the calculation of the probabilities of the form $p_0(u_{\nu_l}|u_\nu)$ or $p_0(u_{\nu_r}|u_\nu)$, e.g., $p_0(u_{R_r}|u_R)$ in \eqref{temp1}. Let us denote any child of the node $\nu$ by $\nu_s$ for generalization. Note that if $u_{\nu}$ and $u_{\nu_s}$ were independent then we would have $p_0(u_{\nu_s}|u_{\nu})=p_0(u_{\nu_s})=\tau$ when $u_{\nu_s}=1$. However, we anticipate a statistical dependency between $u_{\nu}$ and $u_{\nu_s}$ generating a positive covariance. That is, conditioned on the knowledge of $u_{\nu}$, we would like to impose that $u_{\nu_s}$ is more likely to attain the value $u_\nu$ compared to the prior conditions, i.e., $\nu_s$ is likely to inherit the label of its parent. On the other hand, provided that $u_{\nu}$ and $u_{\nu_s}$ are identically dependent, we would have that $p_0(u_{\nu_s}|u_{\nu})=1_{\{u_{\nu}=u_{\nu_s}\}}$, where $1_{\{.\}}$ is the indicator function. To introduce this into the derivations, we parameterize the probability mass function $p_0(u_{\nu_s}|u_{\nu})$ as the weighted average between $p_0(u_{\nu_s})$ and $1_{\{u_{\nu}=u_{\nu_s}\}}$ as
\begin{align}
p_0(u_{\nu_s}|u_{\nu})&=(1-\theta)p_0(u_{\nu_s})+\theta1_{\{u_{\nu}=u_{\nu_s}\}} \label{parameterization}\\
&=(1-\theta)\left(0.5 - u_{\nu_s}(0.5-\tau)\right)+\theta \frac{1+u_{\nu}u_{\nu_s}}{2}, \nonumber
\end{align}
where $\theta \in [0,1]$ is a parameter defining the degree of dependency which generates an increasing covariance as $\theta$ increases in the interval $[0,1]$ such that $\theta=0$ implies the statistical independency of $u_{\nu}$ and $u_{\nu_s}$; and $\theta=1$ implies identical dependency. Then, the probability mass function $p_0(\mathbf u)$ can be calculated using this parametrization based on the recursion in $\eqref{temp1}$. Hence, exhaustively enumerating all possible $\mathbf u$'s and running Algorithm TCS for each of them, one can calculate the false alarm rate $C_\tau$ in \eqref{myeq1}, which is not a practical choice. Instead, through the conditional factorization in \eqref{condindep}, we opt to simplify the expression \eqref{myeq1} and obtain an efficient recursion. To this end, for a given node $\nu$ with depth $1\leq i \leq L-2$, let us define the probability conditioned on $u_\nu$ that Algorithm TCS does not declare a corruption in the tree rooted from $\nu$ denoted by $\mathcal{F}(\nu;u_\nu)$ as
\begin{align*}
\mathcal{F}(\nu;u_\nu)=\sum_{\mathbf u_\nu \in B^{2^{L-i+1}-2}} \mathcal{C}^c((\mathbf u_\nu,u_\nu)) p_0(\mathbf u_\nu|u_\nu).
\end{align*}
Here, $\mathcal{F}(\nu;u_\nu)$ solely depends on the depth variable $i$ due to the symmetric factorization by the conditional independency from parents to children. Therefore, the notation simplifies to $\mathcal{F}(i;1)$ or $\mathcal{F}(i;-1)$. Using the $4$ possible configurations for $(u_\nu=1, u_{\nu_l}, u_{\nu_r})$, we can calculate $\mathcal{F}(i;1)$ as a function of $\mathcal{F}(i+1;.)$. Noting that two of those configurations are the conclusive patterns, termination and corruption patterns, we obtain
\begin{align*}
\mathcal{F}(i;1) = q^2_1(-1)+2q_1(-1)q_1(1)\mathcal{F}(i+1;1)\mathcal{F}(i+1;-1),
\end{align*}
where $q_i(j)=p_0(v_s=j|v=i)$ as a short hand notation; the second term corresponds to the continuation of Algorithm TCS and the first term corresponds to the terminating pattern. Unlike the second term, the first term does not have a multiplier since the search stops at such a node. Note that the corruption pattern is disregarded by definition. Similarly, we also have
\begin{align*}
\mathcal{F}(i;-1)&=q^2_{-1}(1) \mathcal{F}^2(i+1;1) + q^2_{-1}(-1) \mathcal{F}^2(i+1;-1) \nonumber\\
&+2q_{-1}(1)q_{-1}(-1) \mathcal{F}(i+1;1)\mathcal{F}(i+1;-1).
\end{align*}
Recalling that we declare corruptions at leaf nodes on the basis of local anomalies, we can further define
\begin{align*}
\mathcal{F}(L-1;1)=q^2_1(-1) \text{ and }
\mathcal{F}(L-1;-1)=q^2_{-1}(-1),
\end{align*}
and provide the initialization to the recursion $\mathcal{F}(i;1)$ and $\mathcal{F}(i;-1)$. On the other hand, we never declare corruptions at root since we are focused only on localized corruptions, which is an exception and can straightforwardly incorporated in our recursions. In terms of the recursions regarding $\mathcal{F}(i;1)$, the only change is that the corruption pattern should not be disregarded which does not lead to a corruption detection and so does not stop the search. Then, we simply have
\begin{align*}
&\mathcal{F}(0;1) = q^2_1(-1)+q^2_1(1)\mathcal{F}^2(1;1)+\\
&2q_1(-1)q_1(1)\mathcal{F}(i+1;1)\mathcal{F}(i+1;-1);
\end{align*}
and the recursion $\mathcal{F}(i;-1)$ stays valid for $\mathcal{F}(0;-1)$. Now that we have the recursion equations defined for all depth levels on the binary tree, we can efficiently calculate the false alarm rate of Algorithm TCS as follows. Letting $R$ represent the root node, we obtain from \eqref{myeq1}
\begin{align*}
&1-C_\tau = \sum_{\mathbf u \in B^{2^{L+1}-1}} \mathcal{C}^c(\mathbf u) p_0(\mathbf u)\\
&= p_0(u_R=1)\sum_{\mathbf u_R \in B^{2^{L+1}-2}} \mathcal{C}^c((u_R, \mathbf u_R)) p_0(\mathbf u_R|u_R=1)+\\
&p_0(u_R=-1)\sum_{\mathbf u_R \in B^{2^{L+1}-2}} \mathcal{C}^c((u_R, \mathbf u_R)) p_0(\mathbf u_R|u_R=-1).
\end{align*}
Then, recalling that $ p_0(u_R=1)=1-p_0(u_R=-1)=\tau$, the false alarm rate $C_\tau$ is given by
\begin{align}
C_\tau = 1 - \tau \mathcal{F}(0;1) - (1-\tau) \mathcal{F}(0;-1), \label{algFA}
\end{align}
which is equivalent to first calculating the probability that Algorithm TCS never declares a corruption and then subtracting this probability from $1$.

We point out that the false alarm rate $C_\tau$ of Algorithm TCS (Tree-based Corruption Separation) in detecting the corruptions as found in \eqref{algFA} is a data-independent quantity. Therefore, under the simplification through the conditional independency \eqref{condindep}, we conclude that the false alarm rate $\tau$ of the anomaly detection at each node maps to a constant false alarm probability of our corruption detection $C_\tau$. Secondly, even though the dependency parameter $\theta$ does not appear, i.e., hidden, in the expression \eqref{algFA}, $C_\tau$ is clearly affected by $\theta$. For example, if $\theta=1$, i.e., if the binary label of a child node is identically dependent on the parent label and hence $p_0(u_{\nu_s}|u_\nu)=1_{\{u_{\nu_s}=u_\nu\}}$, then it can be shown that $C_\tau=\tau$. If $\theta=0$, i.e., if the binary label of a child node is independent with the parent label and hence $p_0(u_{\nu_s}|u_\nu)=p_0(u_{\nu_s})$, then obviously $C_\tau>\tau$. In Fig. \ref{fig:dagfig}, we plot the hypothetical curves resulted from mapping the constant false alarm rate $\tau$ in detecting the local anomalies to the corruption false alarm rate $C_\tau$ via the described model of conditional independency for several degree of dependencies $\theta$. We experimentally discuss the efficacy of this model in representing the relation between $\tau$ and $C_\tau$ in Section \ref{sec:experiments}. Moreover, the parameter $\theta$ between $u_\nu$ and $u_{\nu_s}$ can also be chosen depth dependent, i.e., $\theta_i$, instead of a uniform choice over the partitioning tree. An example of a depth dependent modeling is given in Section \ref{sec:experiments}. Finally, note that the directed acyclic graph modeling of the anomalous vs normal labeling uniformly holds for all $\alpha$'s in choosing the ranked Euclidean distance. We also discuss the impact of various $\alpha$'s on the fitness of the described dependency structure in Section \ref{sec:experiments}.

In the following section, we explain the important points of our implementation and discuss the corresponding computational complexity.

\section{Computational Complexity} \label{sec:complexity}
Computationally, the main building block in Algorithm TCS (Tree-based Corruption Separation) is the application of the anomaly test defined in \eqref{test}, which computes the train-to-train distance matrix $D_S(i,j)=d(\mathbf s_i, \mathbf s_j)$ and the test-to-train distance vector $D_X(j)=d(\mathbf x, \mathbf s_j)$. Operating on these distances, the score function defined in \eqref{score} for the test instance must be computed, which then requires the computation and sorting of $R_S(\mathbf s_i;K)$. In addition, since we label each node as anomalous or not in our tree expansion, these distances must actually be computed at each node with respect to the corresponding attributes, e.g., $D_{S_{V_\nu}}(i,j)$ and $D_{X_{V_\nu}}(j)$ at a node $\nu$. For this purpose, we adapt the ``integral image" approach in case of using the standard Euclidean distance. Namely, let us define the volume $\mathcal{D}_S(i,j,k)=\sum_{h=1}^{k}({ s_i}_h-{ s_j}_h)^2$, $\forall i,j$ with $1\leq k\leq d$; and $\mathcal{D}_S(i,j,0)=0$, $\forall i,j$ (similarly for $\mathcal{D}_X(j)$ ). Then, we simply have $D_{S_{V_\nu}}(i,j)=\sqrt{\mathcal{D}_S(i,j,k_2)-\mathcal{D}_S(i,j,k_1)}$ at a node $\nu$, where $V_\nu$ corresponds to the set of attributes in positions between $k_1+1$ and $k_2$. The volume $\mathcal{D}_S(i,j,k)$ and the sorting of $R_S(\mathbf s_i;K)$ can be computed offline once the training set is provided, which defines a training phase complexity $O(2^{L+1} N_s^2 \log_2 N_s)$, where sorting is the dominant contributor. For a given test instance, we compute $D_{X_{V_{\nu}}}(j)$ and sort at each node $\nu$ in the expansion of our tree, which defines the test phase complexity $O(2^{L+1}N_s\log_2 N_s)$ for our algorithm, where sorting is the dominant contributor. In case of using the ranked Euclidean distances, since it is not possible to utilize the integral image approach anymore, the computational load is multiplied by constant factors. Next, we illustrate the efficacy of the proposed framework in separating, i.e., detecting and localizing, corruptions and imputing.

\section{Experiments} \label{sec:experiments}
In this section, we test the introduced Algorithm TCS (Tree-based Corruption Separation) over several well-known machine learning data sets subject to synthetically generated data corruptions to demonstrate the performance of the proposed approach. We first discuss the efficacy of the false alarm rate estimation method explained in Section \ref{sec:ACF} in terms of the corruption detection and evaluate the performance of the critical steps in Algorithm TCS, which are the corruption detection, localization and imputation. Then, we report the improvements achieved by the proposed framework in several classification tasks.

\begin{figure}[t]
\centerline{\epsfxsize=7cm \epsfysize=6cm \epsfbox{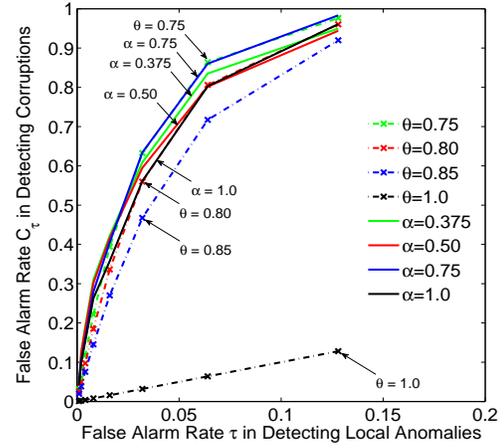}}
\caption{Solid (dash-dot) curves correspond to the realizations (hypothetical results). The constant false alarm rate $\tau$ in detecting the local anomalies maps to a global constant false alarm rate $C_\tau$ in detecting the corruptions with Algorithm TCS (Tree-based Corruption Separation). We observe that setting $\theta\in[0.75,0.8]$ well approximates the relation between $\tau$ and $C_\tau$. In case of the identical dependency, i.e., $\theta=1$, $C_\tau=\tau$.}
\label{fig:dagfig} \vspace{-0.2 in}
\end{figure}

In the first set of experiments, we adapted a $0-1$ digit classification task consisting of a training set of $1500$ samples and a test set of $750$ samples based on the USPS data \cite{uci}. Each of these samples is a $16 \times 16$ gray scale image of either a ``0" image or a ``1" image, where each pixel has a real intensity value in [0,1]. We synthetically generate a corruption as described in Section \ref{sec:PD} and apply to each instance in the test set with probability $\pi=\frac{1}{2}$. To be more precise, for a test instance chosen to be corrupted, we (uniformly) randomly specify a square region of size between $(10-50)\%$ of the total area, i.e., the number of pixels in the chosen region is not less than $25$ and not more than $128$, overwrite each pixel in this region with a value randomly (using the uniform distribution $U_Z(z)$) drawn from the interval $[0,1]$. Then, after the training and test instances are vectorized column wise such that $\mathbf s, \mathbf x \in \mathbb{R}^{256}$, the proposed Algorithm TCS is provided with the clean training data and run over the test set. We emphasize that by this vectorization scheme, the corrupted square region corresponds to multiple corrupted intervals in the vectorized observation. Hence, this example also illustrates that Algorithm TCS can handle multiple corruptions. Ideally, the neighborhood size parameter $K$ for both imputation and corruption separation purposes should be optimized at every node of our binary tree since the data dimensionality from node to node varies. However, we opt not to optimize $K$ for presentational clarity and set as $K=8$ near the midpoint of $[1,16]$, which is empirically found appropriate. Using the $0-1$ digit USPS data, we investigate the response of the Algorithm TCS to the local anomaly detection false alarm rate $\tau \in \Gamma = \{0.001, 0.002, 0.004, 0.008, 0.016, 0.032, 0.064, 0.128 \}$ and the ranked Euclidean distance parameter $\alpha \in \Delta = \{0.375, 0.5, 0.75, 1\}$. As for the depth parameter, we use the deepest possible tree with $L=6$ such that the leaves are associated with $4$ pixels and hence, $1$ pixel at least is then used in the distance calculation with $\alpha=0.375$.
\begin{figure}[t] \vspace{-0.15 in}
\centerline{\epsfxsize=7cm \epsfysize=6.4cm \epsfbox{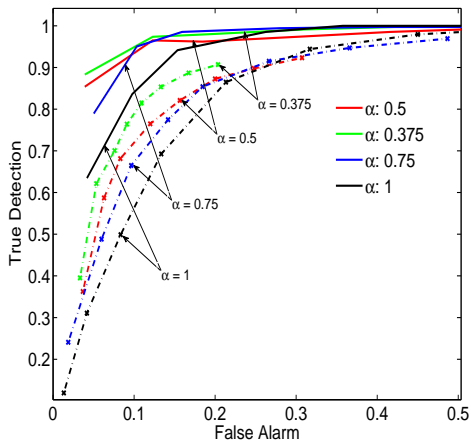}}
\caption{ROC curves for detection and localization of corruptions. Solid (dash-dot) curves correspond to detection (localization) performances. }
\label{fig:ROC} \vspace{-0.2 in}
\end{figure}

In Fig. \ref{fig:dagfig}, we compare the hypothetical false alarm rate $C_\tau$ we derive in Section \ref{sec:ACF} with the corresponding experimental realizations with respect to varying local anomaly detection false alarm $\tau \in \Gamma$. The hypothetical map from $\tau$ to $C_\tau$ is generated with several choices for the dependency parameter, $\theta \in \{0.75, 0.8, 0.85, 1\}$, whereas the realizations correspond to several choices for the distance parameter, $\alpha \in \Delta$. Our experiments indicate that when the statistical dependency $\theta$ in \eqref{parameterization} from a parent node to one of its children nodes, cf. Fig. \ref{fig:dag}, is chosen around $0.75$, the relationship between the local anomaly false alarm rate $\tau$ and the corruption detection false alarm rate $C_\tau$ is accurately modeled. This experimentally shows that the labeling of local anomalies over a binary partitioning tree shown in Fig. \ref{fig:example1} can be considered as a directed acyclic graph. We also observe that in the case of Euclidean distance, i.e., $h_\alpha$ with $\alpha=1$, while $\theta\sim 0.75$ is more accurate for small $\tau$'s, $\theta$ tends to approach $0.8$ as $\tau$ increases for a better modeling. This small deviation mainly happens since the conditional independency assumption explained in Fig. \ref{fig:dag} do not hold in case of Euclidean distance for a certain pattern. Namely, although the labeling for a parent and its children nodes as $-1,1,1$ (a normal parent node with anomalous children nodes) is not possible with the standard Euclidean distance due to the test defined in \eqref{test}, the directed acyclic graph modeling assigns it a positive probability, which then overestimates the corruption detection false alarm rate. Nevertheless, this positive probability is the smallest among the ones assigned to the all possible patterns of three nodes as desired and hence, the ordering of the patterns in terms of their probabilities is still reasonable even in the case of the Euclidean distance. On the contrary, since this pattern is also possible in case of the ranked Euclidean distance, the accuracy of our hypothetical results improves as $\alpha$ decreases.
\begin{figure}[t]
\centerline{\epsfxsize=7cm \epsfysize=6.1cm \epsfbox{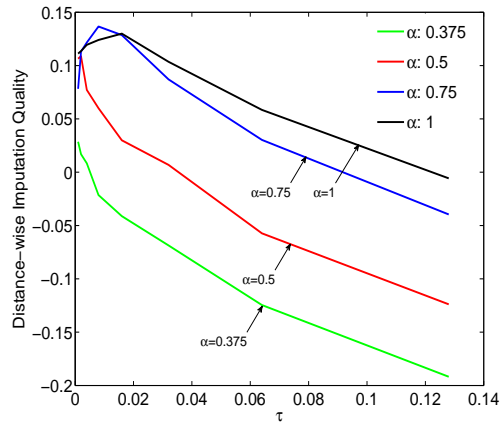}}
\caption{Distance-wise imputation quality with $\tau\in\Gamma$.}
\label{fig:impqual} \vspace{-0.2 in}
\end{figure}

Next, we study the corruption detection and localization performance of our algorithm on the $0-1$ digit USPS data. In Fig. \ref{fig:ROC}, we plot the empirical false alarm rates versus the empirical true detection rates in terms of both corruption detection and corruption localization with respect to $\tau \in \Gamma$. Here, the true detection rate is the empirical probability, i.e., relative frequency, of a truly corrupted data instance (data attribute in case of localization) to be declared corrupted and, the false alarm rate is the empirical probability of a truly uncorrupted data instance (data attribute in case of localization) to be declared corrupted. As we discuss in Section \ref{sec:thethemethod}, the ranked Euclidean distance is experimentally shown to produce a better detection as well as localization performance on the USPS data as $\alpha$ decreases. Recall that for a small $\alpha$ around $0.5$, we enforce a corruption to be widely spread for Algorithm TCS to detect it at a node, which then clearly improves the localization. Similarly, the corruption detection performance also improves as $\alpha$ decreases. Since the ranked Euclidean distance disregards a certain fraction of largest attribute-wise deviations, Algorithm TCS behaves conservatively in declaring corruptions. This reduces the false alarms in terms of the local anomalies and in turn, reduces the false encounters of the terminating pattern shown in Fig. \ref{fig: simplecorruption}b. Hence, the corruption search is not stopped mistakenly, and Algorithm TCS does not miss certain corruptions, which leads to a better detection rate with the ranked Euclidean distance using a small $\alpha$ around $0.5$. We emphasize that the local anomaly detection false alarm rate $\tau$ can be set independently for detection and localization to precisely determine the operating point on the ROC curves in Fig. \ref{fig:ROC}. However, in this study, we use one single $\tau$ in all phases of Algorithm TCS. Note that when the false alarm rate is set around $0.1-0.2$, our algorithm is able to provide a detection rate around $0.9$ and a localization rate around $0.8$.
\begin{figure}[t]
\centerline{\epsfxsize=9.0cm \epsfysize=5.0cm \epsfbox{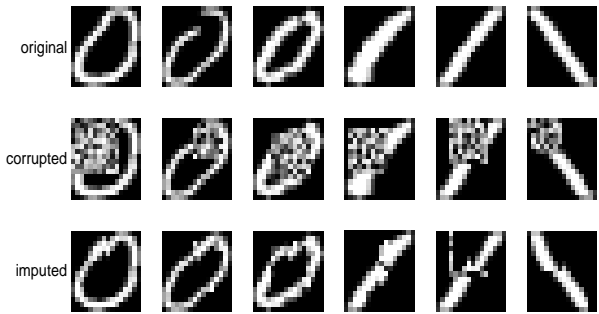}}\vspace{-0.2 in}
\caption{Several visual examples on USPS dataset.}
\label{fig:visualexamples} \vspace{-0.2 in}
\end{figure}

On the other hand, the ranked Euclidean distance parameter $\alpha$ cannot be made arbitrarily small since, as $\alpha$ decreases, the determination of the neighbors of a test instance in the reference set degrades and hence, the imputation quality is adversely affected. Observe that with small $\alpha$, only a small fraction of attributes are used in determination of $\mathcal{N}_{N_s}(\mathbf x_{V_{\nu_s}})$ in \eqref{empiricalneighborhood} despite that the rest of the attributes might be informative through the local correlations and hence, the imputation quality degrades. We illustrate this effect in Fig. \ref{fig:impqual}, where we use the improvements in the distance wise deviations after imputation to measure the imputation quality. For this purpose, we define
\begin{align}
\frac{1}{N_c}\sum_{i=1}^{N_c} \frac{ h_{\alpha=1}(\bar{\mathbf x}_i,\mathbf x_i)-h_{\alpha=1}(\bar{\mathbf x}_i,\hat{\mathbf x}_i) }{h_{\alpha=1}(\bar{\mathbf x}_i,\mathbf x_i)} \label{distimpqual}
\end{align}
as the distance wise imputation quality, where $N_c$ is the number of the corrupted test instance (which is approximately $750\pi$), $\mathbf x_i$ is a corrupted test instance, $\bar{\mathbf x}_i$ is the uncorrupted original instance and $\hat{\mathbf x}_i$ is the corresponding instance after imputation. Note that this quality metric measures on average that how much of the distortion by the corruption is recovered after imputation. The average imputation quality defined in \eqref{distimpqual} is plotted versus the local anomaly detection false alarm $\tau \in \Gamma$ in Fig. \ref{fig:impqual}. We first observe that for large $\tau$, since the false alarm rate is also large, the imputation even further disturbs the data. Secondly, for small $\tau$ around $0.01$, the proposed imputation technique is able to correct a corrupted instance up to $12 \%$ in case of $\alpha \sim 0.75$. Moreover, our experiments also indicate that for $\alpha$ less than $0.5$, the ranked Euclidean distance is not able to produce desirable results despite its superiority in terms of detection and localization, which reinforces our discussion about that $\alpha$ cannot be made too small.

Unlike an MMSE based approach, our MAP based imputation does not target minimizing the reconstruction error but the most likely replacement for a corruption. Therefore, the distance-wise imputation quality measure in \eqref{distimpqual} is not fair to use for a comparison. Indeed, an MMSE based estimator for imputation would produce visually blurry results, for instance on the USPS data. In this regard, we present several visual examples that the proposed framework generates on the USPS data with $\tau = 0.016\text{, }\alpha=0.75 \text{, }K=8$ in Fig. \ref{fig:visualexamples}. Note that the presented visual examples tend to generate image gradients that are naturally aligned with the image statistics, since our imputation method is not based on an averaging to minimize the reconstruction error but filling in the missing part with the most likely candidate extracted from the reference data. We also observe some cases, where the corruption along a border between the cells of our partitioning tree remains after the imputation, cf. the second last column in Fig. \ref{fig:visualexamples}. The residual corruptions in such cases can be handled by increasing the depth of the tree or using m-ary trees as opposed to binary splits, which is not in the scope of this study.
\begin{figure}[t]
\centerline{\epsfxsize=9cm \epsfysize=4cm \epsfbox{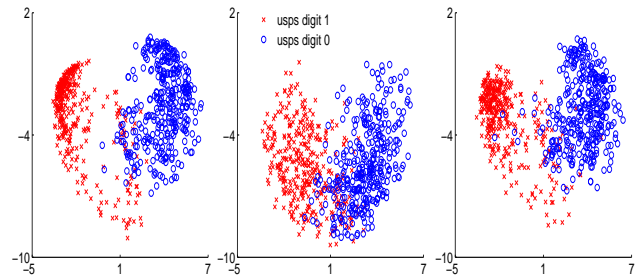}}
\caption{On the left is the (uncorrupted) true data scatter; mean separation between two classes: $5.95$, linear SVM accuracy: $99.71 \%$. In the middle is the corrupted data scatter; mean separation: $4.19$, classification accuracy: $90.57 \%$. On the right is the imputed data scatter; mean separation: $5.37$, classification accuracy: $96.85 \%$, which corresponds to $\sim 68.0\%$ improvement both in terms of the mean separation and classification.}
\label{fig:thesep} \vspace{-0.2 in}
\end{figure}

In addition to the visual comparisons, we also evaluate the performance of the introduced framework in terms of the classification purposes. On the described $0-1$ digit USPS data, we report the data scatter plots of the test instances in Fig. \ref{fig:thesep}, where we project the original, corrupted and imputed test data onto the two eigen vectors of the training set with the largest eigenvalues for visualization. We clearly observe a better class separation between two classes after the imputation, when compared to the class separation in the corrupted data. The distance between the class means is calculated to be $5.95$ in the original uncorrupted data, $4.19$ in the corrupted data and $5.37$ after the imputation. This approximately corresponds to a $68\%$ performance improvement achieved by the proposed framework both in terms of the mean separation and classification accuracy, cf. Fig. \ref{fig:thesep}.
\begin{table*}[t]
\caption{Improvements in terms of the classification accuracies achieved by the proposed framework. \vspace{-0.1in}} 
\centering 
\begin{tabular}{c c c} 
\hline
Data Sets (dimensionality) & Uncorrupted Training & $\% 5$ Corrupted Training \\
Avg.  $\mathbf a \mathbf c \mathbf c(\bar{\mathbf X})$ / $\mathbf a \mathbf c \mathbf c(\mathbf X)$ & Avg. $\%$ Improvements with its Std. below & Avg. $\%$ Improvements with its Std. below\\
 & TCS-MAP / TCS-NN / 4-NN (16-NN) & TCS-MAP / TCS-NN / 4-NN (16-NN)\\ [0.5ex] 
\hline \vspace{-0.1in}
& & \\
Image (18) & 55.60 / {\bf59.14} / 42.80 (10.34)  & {\bf40.96} / 40.95 / 27.84 (9.34) \\ \vspace{0.05in}
84.62 / 57.16         & 1.81 / 2.58 / 3.20 (1.47)  & 1.32 / 1.92 / 3.32 (1.79) \\
Ringnorm (21) & {\bf46.85} / 46.15 / 31.99 (19.42)  & {\bf43.36} / 39.60 / 26.36 (18.35) \\ \vspace{0.05in}
76.90 / 64.40         & 4.23 / 3.27 / 5.40 (4.67)  & 4.25 / 4.89 / 4.21 (5.62) \\
Twonorm (21) & 31.16 / {\bf31.36} / 14.75 (16.46)  & {\bf22.60} / 20.97 / 11.64 (12.51) \\ \vspace{0.05in}
96.46 / 85.84         & 3.51 / 1.91 / 2.09 (2.71)  & 3.49 / 1.96 / 1.70 (3.55) \\
Svmguide3 (21) & 62.33 / {\bf68.86} / 60.57 (6.86)  & 56.11 / {\bf59.80} / 43.03 (12.04) \\ \vspace{0.05in}
81.08 / 68.74         & 3.03 / 3.29 / 2.09 (5.38)  & 2.79 / 3.71 / 3.62 (4.15) \\
Breast Cancer (30)& 72.19 / {\bf72.24} / 19.07 (46.08) & {\bf57.74} / 55.07 / 18.80 (42.33)\\ \vspace{0.05in}
97.30 / 85.66         & 3.65 / 3.47 / 2.99 (6.07)  & 2.93 / 3.11 / 5.34 (5.68) \\
Spam Data (57)& {\bf77.38} / 77.38 / 53.86 (64.85) & {\bf69.46} / 68.18 / 33.78 (35.43)\\ \vspace{0.05in}
88.66 / 62.56         & 1.69 / 1.71 / 1.43 (1.58)  & 2.01 / 1.88 / 2.07 (2.31) \\
Sonar (60)& 40.25 / {\bf48.00} / -25.83 (-38.20) & 40.33 / {\bf44.96} / -47.86 (-27.45)\\ \vspace{0.05in}
75.51 / 70.43         & 28.42 / 26.68 / 16.87 (36.62)  & 24.64 / 25.79 / 25.39 (36.74) \\
BCI (117) & 27.75 / {\bf34.12} / 31.10 (8.19)  & 15.12 / {\bf15.47} / 11.76 (8.79) \\ \vspace{0.05in}
79.55 / 60.08         & 7.75 / 6.60 / 5.79 (9.33)  & 5.30 / 6.26 / 3.20 (5.16) \\
Digit1 (241) & 76.03 / {\bf81.96} / 39.29 (40.06)  & 25.29 / {\bf27.36} / 19.49 (17.70) \\ \vspace{0.05in}
95.96 / 87.80         & 4.61 / 3.30 / 2.44 (4.16)  & 3.65 / 2.92 / 3.57 (3.28) \\
G241c (241) & {\bf13.04} / 12.72 / -13.41 (32.45)  & {\bf9.54} / 9.46 / 3.86 (10.95) \\ \vspace{0.05in}
87.42 / 76.00         & 7.65 / 8.13 / 5.79 (5.77)  & 3.34 / 2.79 / 1.61 (3.14) \\
G241n (241) & {\bf18.34} / 16.55 / -18.93 (34.64)  & {\bf4.93} / 2.85 / 1.47 (6.89) \\ \vspace{0.05in}
85.60 / 77.54         & 7.07 / 7.54 / 8.16 (5.40)  & 3.25 / 3.59 / 3.16 (4.49) \\ [-0.25ex]
\hline
\end{tabular}
\label{table:nonlin} \vspace{-0.25in} 
\end{table*}

We emphasize that this study is the first to jointly handle localized data corruptions in one complete framework consisting of detection, localization and imputation phases, which are designed jointly and completely model free for the same goal of separating a possible corruption from a test instance and imputing the affected data attributes. In this respect, the proposed algorithm is a comprehensive one such that it operationally covers the partial solutions in the corresponding literature. For this reason, our algorithm is not comparable to any of those partial solutions and it is not possible to provide a fair comparative analysis. Nevertheless, we compare the proposed framework with a baseline of algorithms constructed using the methods \cite{mmethod1, mmethod2} in terms of the classification tasks over the several well-known machine learning data sets \cite{uci, clayton}. One of these methods, ``TCS-NN", consists of the same tree-based corruption separation (TCS) procedure that we propose but utilizes -instead of our MAP imputation- the nearest neighbor (NN) imputation technique \cite{mmethod1}, which finds the nearest neighbor of a corrupted instance with respect to the sibling attributes of the corrupted node and imputes. The other method, ``M-NN", also utilizes the nearest neighbor imputation but does not have a fine/detailed corruption separation step. Instead, it splits an instance into $M$ different segments \cite{mmethod2}, applies anomaly detection to each segment and imputes an anomalous segment by its nearest neighbor that is found with respect to the neighboring segment.

In these experiments, we use a depth-$4$ tree for our Algorithm TCS leading to $16$ leaves/segments in the finest level with $K=8$. For each data set, after scaling each data attribute into the interval [0,1], we randomly choose $11$ splits of the scaled data such that $2/3$ of each split is reserved as the training (reference) data set (at most $1000$ instances), and the rest is reserved as the test data set (at most $500$ instances) in each split. Moreover, every instance in the test set of each split is randomly corrupted/overwritten from the uniform distribution with the support $[0,1]$ in a random interval of attributes, which includes at least $10 \%$ of the attributes (dimensionality) and at most $50 \%$ of them. Since $30=\frac{50+10}{2} \%$ of each test instance is corrupted on average, choosing $M=4$ is appropriate for the method ``M-NN". We also present results for the case $M=16$. The first split is used for parameter selection purposes\footnote{The same values for the parameters $\alpha,\tau$ is used for both ``TCS-MAP", and ``TCS-NN" to fairly and clearly observe the effect of using NN imputation instead of MAP imputation since using the same rate $\tau$ for both methods leads to the same CFAR in corruption detection. In another separate experiment cf. Table \ref{table:impute}, we directly and explicitly compare the two imputation methods with the standard Euclidean distance only. The same $\tau$ is also used for ``M-NN", which definitely favors ``M-NN" since it corresponds to a lower CFAR for ``M-NN". Euclidean distance, $\alpha=1$, is always used for ``M-NN". Depth is always $4$ with $K=8$. $C$ is always common to all methods.}, e.g., $C$ parameter of a linear SVM, $\alpha$, and the remaining $10$ splits are used for performance analysis. Then, in each of the remaining $10$ splits, we train a linear SVM classifier on the training set and compute the classification accuracy on the uncorrupted, corrupted and the imputed test data. In Table \ref{table:nonlin}, we report the average of improvements and the corresponding standard deviation of the average improvement (std. of the improvements divided by $\sqrt{10}$ is reported, i.e., std. of the mean value estimator) in terms of the classification accuracy on the imputed test data over $10$ splits. The improvement in classification accuracy is calculated as
\begin{align}
\text{$\%$ Improvement } = 100 \times \frac{\mathbf a \mathbf c \mathbf c(\hat{\mathbf X})-\mathbf a \mathbf c \mathbf c(\mathbf X)}{\mathbf a \mathbf c \mathbf c(\bar{\mathbf X})-\mathbf a \mathbf c \mathbf c(\mathbf X)}, \label{accimpdef}
\end{align}
where $\mathbf a \mathbf c \mathbf c(\bar{\mathbf X})$ ($\mathbf a \mathbf c \mathbf c(\mathbf X$), $\mathbf a \mathbf c \mathbf c(\hat{\mathbf X}$)) is the classification accuracy -after clean data training- on the original uncorrupted test data $\bar{\mathbf X}$ (on the corrupted test data $\mathbf X$, on the imputed data $\hat{\mathbf X}$). Furthermore, in order to evaluate the robustness to corruptions in the training sets and to address the cases where clean data is not available, we randomly choose the $5 \%$ of each training data of each split, corrupt with the same corruption model and then repeat\footnote{Exactly the same splits with the same parameters are used. The only difference is that $5 \%$ of each training set is corrupted. Note that in these experiments, we first train a linear SVM using clean training data and then test the trained model on uncorrupted (clean) test data, corrupted test data and corrupted test data imputed by the clean training data and corrupted test data imputed by $\% 5$ corrupted training data.} the same experiments. The results are summarized in Table \ref{table:nonlin}.

We observe that the proposed framework is significantly successful at undoing the adverse effects of corruption and always provides (positive) improvements up to $\sim 80 \%$ after imputation in terms of the classification performance. Additionally, when the accuracy drop due to the corruptions is relatively low, e.g., $5$ units of drop from $80 \%$ to $75 \%$, even a $1$ unit gain after imputation corresponds to $20 \% = \frac{1}{5}$ improvements, which naturally results in relatively high standard deviations, cf. Sonar data set in Table \ref{table:nonlin}. Moreover, the proposed framework is shown to be robust to corruptions in the training data by these experiments.
\begin{table}[t]
\caption{Performance of the MAP Imputation. \vspace{-0.1in}} 
\centering 
\begin{tabular}{c c c c c c c} 
\hline \vspace{-0.1in}
& & & &   \\
$K $ & $1$ & $4$& $8$ & $12$& $16$   \\
MSE & $2.90$& $2.31$& $2.19$& $2.14$&   $2.12$   \\
$\%$ Classification Accuracy &  $79.49$& $80.13$&  $80.45$& $80.74$&   $80.86$ \\ [-0.25ex] 
\hline
\end{tabular}
\label{table:impute} \vspace{-0.25in} 
\end{table}

Our algorithms, ``TCS-MAP" (Tree-based Corruption Separation with MAP imputation), perform significantly better than the method ``M-NN" and comparably with the method ``TCS-NN". We point out that the method ``TCS-NN" strongly relies on the proposed Algorithm TCS, which is the reason underlying the success of ``TCS-NN" since the method ``M-NN" is outperformed by ``TCS-NN" (Both of the methods ``M-NN" and ``TCS-NN" uses the NN imputation but ``M-NN" does not have the proposed tree-based corruption separation step). The method ``M-NN" occasionally even further corrupts the data, cf. the negative improvements for Sonar or G241c or G241n. Moreover, it is difficult to choose between the methods ``4-NN" and ``16-NN" since there is no clear superiority. Namely, there is definitely a scaling issue for the method ``M-NN". If the corruption in an instance covers a small portion, e.g., $10 \%$, then choosing $M$ too large would leave several undetectable corruptions in addition to imputing large chunks of clean data due to the false alarms (negative improvements). Similarly, if the corruption in an instance covers a large portion, e.g., $50 \%$, then choosing $M$ too small would not only again harden the anomaly detection (due to the use of insufficient data in detecting corruptions) but also complicate the imputation since on what to condition the imputation becomes ambiguous. For a good imputation in this case, one would need to identify (with significant imperfections due the use of insufficient data because of a small $M$) all the corrupted and uncorrupted segments; and then condition the imputation on uncorrupted ones, which leads to a non-homogenous different evaluation for every instance that requires computationally a very high load. Therefore, choosing an appropriate $M$ is in general hard since it must depend on the amount of the corruption, which might be unknown and random. The proposed framework resolves this scaling issue in a computationally efficient way via the binary searches and fast imputations. Moreover, our algorithm ``TCS-MAP" is experimentally shown to be also robust to corruptions in training data in the sense that it strongly preserves its corruption separation/imputation capabilities even after including $5 \%$ corruptions in training.

Lastly, our estimator that is used for imputation asymptotically (as the data size increases) recovers the true MAP estimator; and the NN estimator is certainly asymptotically sub-optimal both in the mean square error (MSE) and the likelihood maximization sense. However, the MAP imputation and the NN imputation (only when combined with the proposed tree based corruption separation) performs comparably in our experiments, which is due to the sparsity of the data compared to the dimensionality. We first note that the MAP estimator yields the NN estimator, if the neighborhood size is set $K=1$ in the imputation phase, which can easily be achieved via cross validation for $K$. Clearly, with such a cross validation, the MAP imputation can only perform better than the NN imputation and we opt not to optimize $K$ for presentational clarity. Additionally, the NN imputation is definitely sensitive to corruptions in the training data since the attributes of the nearest neighbor that are used for imputation can also be corrupted with a certain probability, e.g., with probability $0.05$ in our experiments. On the other hand, that possibly (with probability $0.05$) corrupted nearest neighbor would achieve a lower score value $\hat{p}_K(x)$ if it was truly corrupted and it would not be picked by our MAP estimator for imputation. Thus, the proposed MAP estimator can handle such situations and is robust to corruptions, where the NN imputation performance potentially degrades more. In fact, in our experiments, the MAP imputation either enhances its superiority or becomes superior or approaches NN imputation for most of the data sets after including corruptions in training, e.g., Image or Ringnorm data sets in Table \ref{table:nonlin}.

To further demonstrate the power of the proposed estimator, we devise a separate experiment, where we use a data set consisting of two Gaussian components with unitary covariances. We use the means $[1,-1]$ and $[-1,1]$ for positive and negative classes, respectively. We generate $1000$ samples as the training data, ($500$ for each class), and next suppose that the second attribute of each sample is corrupted/missing; and therefore imputed by our MAP estimator with varying neighborhood size $K$ and the NN estimator. Note that in this part, we use standard Euclidean distance only. After repeating this $100$ times, the resulting imputed data is compared to the original training data in the MSE sense and in terms of the classification accuracy. We summarize our findings in Table \ref{table:impute}, where the MAP imputation with $K=1$ coincides with the NN imputation. The MAP imputation is consistently better than the NN imputation as expected. For instance, when $K=12$, we obtain $26 \%$ improvements in the MSE sense; and $10 \%$ improvements in terms of the classification (original classification accuracy is around $92 \%$).

\section{Conclusion} \label{sec:conclusion}
In this paper, we proposed a comprehensive framework for handling
localized and severe data corruptions. The novel contributions of the proposed framework
includes (i) a first algorithm to jointly detect and localize such corruptions by identifying the local anomalies, (ii) a \emph{Maximum A Posteriori} based
estimator for imputation; and a distance measure for corruption separation purposes, (iii) computational efficiency via the binary searches and the fast imputations, and (iv) a characterization for anomalous observations, e.g., rarities, incompatible combinations and corruptions.
We point out that our algorithm does not assume prior information or a model for the input data and instead, works in a completely data
driven way. Furthermore, we conducted a false alarm rate analysis and showed that
the desired false alarm rate in detecting corruptions can be set independently with the input data.
Our algorithm is tested against the synthetically generated corruptions in several well-known machine learning data sets and experimentally shown
to provide signiﬁcant improvements in terms of classiﬁcation purposes with strong corruption separation capabilities. The proposed algorithms outperform the typical approaches and are robust to varying training phase conditions.
\ifCLASSOPTIONcaptionsoff
  \newpage
\fi
\bibliographystyle{IEEEbib}

\end{document}